\documentclass[runningheads]{llncs}

\usepackage[T1]{fontenc}
\usepackage{url}

\usepackage{graphicx}
\usepackage{booktabs}   
\usepackage{caption}    
\usepackage{siunitx}    
\usepackage{adjustbox}

\begin{document}

\title{Increasing AI Explainability by LLM Driven Standard Processes}

\author{Marc Jansen \and Marcel Pehlke}

\authorrunning{M. Jansen et al.}

\institute{Computer Science Institute\\University of Applied Sciences Ruhr West\\Bottrop, Germany\\
\email{\{marc.jansen,marcel.pehlke\}@hs-ruhrwest.de}}

\maketitle              

\begin{abstract}
This paper introduces a novel approach to increasing the explainability of artificial intelligence (AI) systems by embedding Large Language Models (LLMs) within standardized analytical processes. While traditional explainable AI (XAI) methods focus on feature attribution or post-hoc interpretation, the proposed framework integrates LLMs into well-defined decision models such as Question–Option–Criteria (QOC), Sensitivity Analysis, Game Theory, and Risk Management and the like. By situating LLM reasoning within these formalized structures, the approach transforms opaque inference into transparent, auditable decision traces. A layered architecture is presented that separates the unexplainable reasoning space of the LLM from the explainable process space above it. Empirical evaluations demonstrate that the system can reproduce human-level decision logic in decentralized governance, systems analysis, and strategic reasoning contexts. The results indicate that LLM-driven standard processes provide a promising foundation for reliable, interpretable, and verifiable AI-supported decision-making.

\keywords{Explainable AI, Large Language Models, Standard Processes}
\end{abstract}

\section{Introduction}
Artificial Intelligence (AI) has achieved remarkable advances in recent years, particularly through deep learning and large-scale generative models. Yet, these systems frequently operate as opaque black boxes, making their internal reasoning inaccessible to human understanding. This lack of transparency poses significant challenges to accountability, reliability, and ethical governance, especially when AI is deployed in critical decision-making domains such as finance, healthcare, or policy. The emerging field of Explainable Artificial Intelligence (XAI) seeks to address this gap by developing techniques that make model behavior interpretable and trustworthy to human stakeholders.

While existing XAI methods, such as feature attribution, saliency mapping, and surrogate modeling, have contributed valuable insights, they often remain limited in scope. Many of these approaches describe what influenced a model’s decision, but not how the reasoning process unfolded. Moreover, their results are often highly technical, requiring expert interpretation rather than offering explanations accessible to general users or decision-makers. As a result, the promise of explainability is frequently constrained by the gap between mathematical interpretability and human-centered understanding.

This paper proposes a complementary perspective: instead of merely explaining opaque models post hoc, AI systems can be structured to reason through inherently explainable processes. We introduce the concept of LLM-driven standard processes, in which Large Language Models (LLMs) execute well-established analytical frameworks that are transparent by design, such as Question\-–Option\-–Criteria (QOC) analysis, Sensitivity Analysis, Game Theory, and Risk Management. These processes, long used in human decision-making and systems analysis, possess clear structures, traceable reasoning steps, and interpretable outputs. Embedding LLM reasoning within such frameworks allows the system to generate not only an answer but also a documented rationale showing how that answer was reached.

To operationalize this concept, we present a layered architecture that separates the unexplainable reasoning space of the LLM from the explainable process space of standardized analytical models. The lower layer leverages the generative and interpretive capabilities of the LLM to supply contextual reasoning, while the upper layer enforces formal structures, statistical validation, and auditability. This design introduces an Explainability Barrier that clearly delineates where reasoning becomes measurable, interpretable, and accountable.

We demonstrate the feasibility of this architecture through several prototype implementations and empirical evaluations. Specifically, we apply the LLM-driven QOC framework to decentralized investment decisions in a real-world DAO context, replicate a Vester-style sensitivity analysis on logistics data, and model strategic crisis dynamics using game-theoretic reasoning. Across these evaluations, the proposed system reproduces key elements of human-level reasoning while maintaining transparent analytical structures and quantifiable results.

Overall, this work contributes to the field of explainable AI by showing that LLMs can act not merely as reasoning engines but as explainability enablers when embedded within formalized decision processes. The approach establishes a pathway toward AI systems that are both powerful and transparent, capable of complex reasoning while producing outputs that remain interpretable, auditable, and aligned with human standards of rational explanation.

\section{State of the Art}
Modern AI systems, particularly deep learning models, often act as black boxes whose decision logic is opaque. This opacity limits user trust and hinders adoption in critical domains\cite{xai_llm}. The field of Explainable AI (XAI) therefore seeks to make AI behavior more interpretable and accountable. Existing approaches can be divided into intrinsically interpretable models and post-hoc explanation methods, which explain the outputs of complex models without modifying their internal mechanisms\cite{post-hoc-xai}.

\subsection{Explainable AI}
Post-hoc XAI techniques such as LIME and SHAP approximate local model behavior or assign Shapley values to quantify each feature’s contribution to a prediction\cite{lime,shap-1,shap-2}. These methods have improved transparency in areas like medicine or finance, but they often remain technical and difficult for non-expert users to interpret. Other analytical tools, such as sensitivity analysis, help assess how input variations influence model outputs\cite{xai_sensa}. Together, these methods contribute to fairness, accountability, and regulatory compliance (e.g., GDPR)\cite{xai_llm}.

\subsection{Large Language Models for Explainability}
Despite such progress, current XAI methods still struggle to explain how decisions are reached. Large Language Models (LLMs) offer a new paradigm: their ability to generate coherent natural language allows them to translate complex model outputs into human-understandable narratives. Moreover, chain-of-thought (CoT) prompting enables models to make their reasoning steps explicit, increasing transparency and even improving accuracy\cite{xai_llm,xai_llm_arch}. Studies show that structured reasoning and fine-tuning can raise model performance from around 60\% to over 90\% on complex decision tasks, demonstrating that explainability and reliability can reinforce each other\cite{xai_llm_arch}.

\subsection{LLM-Driven Structured Decision Processes}
A promising next step is to embed LLMs within structured analytical processes that inherently yield explanations. Frameworks like Question–Option–Criteria (QOC) formalize decision-making through transparent evaluation tables that naturally document rationale\cite{qoc}. Recent experimental systems use LLM agents as stakeholders performing QOC analyses collaboratively, producing both recommendations and their justifications. Similarly, multi-agent LLM frameworks simulate deliberation or negotiation, enhancing interpretability through debate-like reasoning\cite{common_ground}.

In parallel, game-theoretic and sensitivity-analysis models are being extended with LLMs to explore strategic behavior, scenario generation, and systemic robustness\cite{llm_game_theory,llm_sensa,llm_banks}. These approaches suggest that LLM-driven analytical structures can transform opaque reasoning into traceable, auditable processes—laying the foundation for explainable, process-based AI systems.

Modern AI systems, particularly deep learning models, often act as black boxes whose decision logic is opaque. This opacity limits user trust and hinders adoption in critical domains\cite{xai_llm}. The field of Explainable AI (XAI) therefore seeks to make AI behavior more interpretable and accountable. Existing approaches can be divided into intrinsically interpretable models and post-hoc explanation methods, which explain the outputs of complex models without modifying their internal mechanisms\cite{post-hoc-xai}.

\section{Architecture}
\begin{figure}[htb]
\centering
\includegraphics[width=0.9\textwidth]{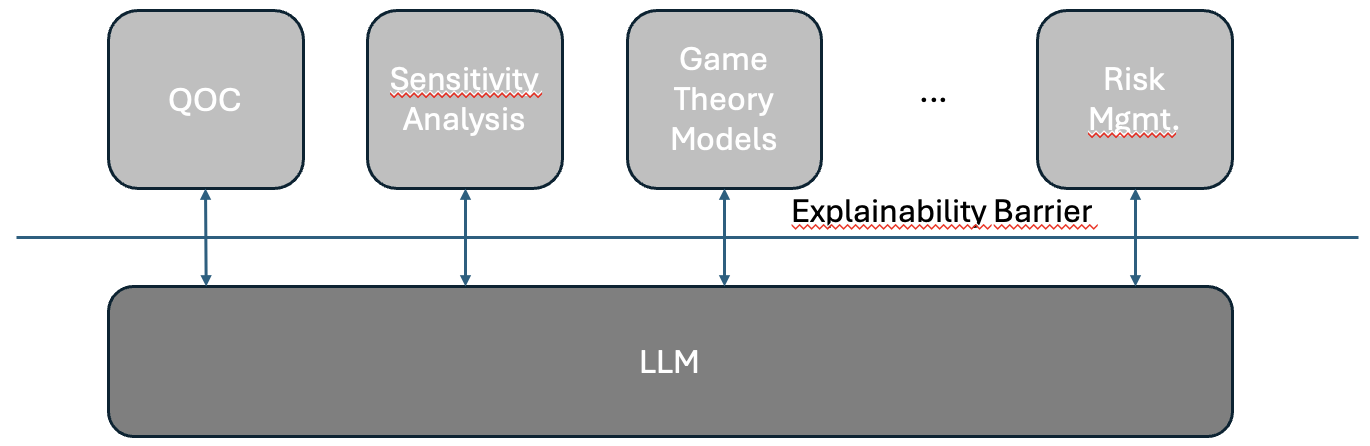}
\caption{Layered architecture of LLM-driven standard processes and the explainability barrier.}
\label{fig:architecture}
\end{figure}

The proposed architecture introduces a layered framework that combines the generative reasoning capabilities of Large Language Models (LLMs) with standardized, transparent analytical processes positioned on top of them. As illustrated in Figure \ref{fig:architecture}, this design establishes a conceptual separation between the unexplainable reasoning space of the LLM and the explainable reasoning space of formalized processes such as for example Question–Option–Criteria (QOC), Sensitivity Analysis, Game Theory Models, and Risk Management.

\subsection{Layered Structure and Explainability Barrier}
At the core of the system lies the LLM, which acts as a general-purpose inference engine capable of interpreting natural language, generating reasoning steps, and providing contextually rich answers. However, despite its remarkable performance, the internal mechanisms of an LLM remain largely opaque: its reasoning paths and parameter activations are not interpretable in human-understandable terms.

To overcome this limitation, a layer of LLM-driven standard processes is placed above the model. Each of these processes defines a formal structure for decision-making or analysis, such as options and criteria in QOC, variables and influence matrices in Sensitivity Analysis, or payoffs and equilibria in Game Theory. Within these frameworks, the LLM’s role is restricted to performing well-defined sub-tasks, such as identifying relevant inputs, generating candidate options, or estimating relationships. The process logic itself, however, remains fully deterministic and analyzable, providing a clear interface for evaluating and understanding the outcomes.

The boundary between the two layers is conceptualized as the Explainability Barrier. Below this barrier lies the LLM’s black-box reasoning, while above it reside structured analytical processes that expose interpretable, measurable, and auditable outputs. By systematically transferring more decision logic from below to above the barrier, the architecture increases the proportion of explainable reasoning in the overall system. This gradual shift is a central goal of the approach: to minimize the influence of opaque model behavior and maximize traceability through formalized process design.

\subsection{Function of the Upper Layer: Standardized Analytical Processes}
The upper layer hosts multiple independent yet compatible analytical modules, for example:
\begin{itemize}
\item QOC (Question–Option–Criteria): for evaluating decision alternatives against weighted criteria.
\item Sensitivity Analysis: for quantifying systemic interdependencies and influence strengths.
\item Game Theory Models: for analyzing strategic interactions, equilibria, and trade-offs.
\item Risk Management: for identifying, consolidating, and evaluating project risks.
\end{itemize}
Each of these modules defines a structured reasoning schema that can be populated and executed by LLM-based agents. Because the schemas are mathematically defined, using weights, scores, matrices, or payoff functions, their results can later be subjected to, e.g., statistical validation, variance analysis, or outlier detection. This enables transparency in terms of why a specific decision or outcome was produced, even if the internal reasoning of the LLM agents remains hidden.

\subsection{Expanding the Explainable Space}
A major design goal of this architecture is to progressively extend the explainable space upward, reducing reliance on unstructured LLM reasoning. This can be achieved by formalizing more process steps, such as option generation, weighting, or evaluation, into transparent mathematical and logical operations. Over time, as additional standardized models (e.g., scenario planning, system dynamics, or causal reasoning) are integrated into the upper layer, the Explainability Barrier moves downward, effectively isolating the black-box elements of the LLM while retaining their creative and interpretative benefits.

In summary, the architecture represents a hybrid explainability paradigm: it combines the linguistic and contextual strengths of LLMs with the interpretability and analytical rigor of established decision-support frameworks. The result is a system in which unexplainable intelligence serves explainable processes, enabling reproducible, transparent, and auditable AI-driven decision support.

\section{Implementation}
Established analytical frameworks can be reimagined through LLM-based agent systems to enhance transparency and consistency in complex decision-making. The following subsections illustrate how classical methods: QOC, Sensitivity Analysis, Game Theory, and Risk Management; can be transformed into LLM-driven standard processes, combining formal analytical rigor with the contextual reasoning capabilities of modern AI.

\subsection{QOC - Question Option Criteria}
The Question–Option–Criteria (QOC) framework originates from the domain of design space analysis \cite{qoc}, yet its underlying principles are broadly applicable to a wide range of decision-making contexts. The core idea is to decompose a decision problem into three elements: a question to be addressed, a set of options representing potential answers, and a collection of criteria that define what aspects are relevant for evaluating these options. Each criterion is assigned a weight to reflect its relative importance within the decision context. Subsequently, each option is assessed with respect to each criterion, indicating how well the option satisfies that particular aspect. By aggregating these assessments, typically as the weighted sum of the option–criterion evaluations, the most favorable option can be identified as the one achieving the highest overall score.

\begin{figure}[htb]
\centering
\includegraphics[width=0.9\textwidth]{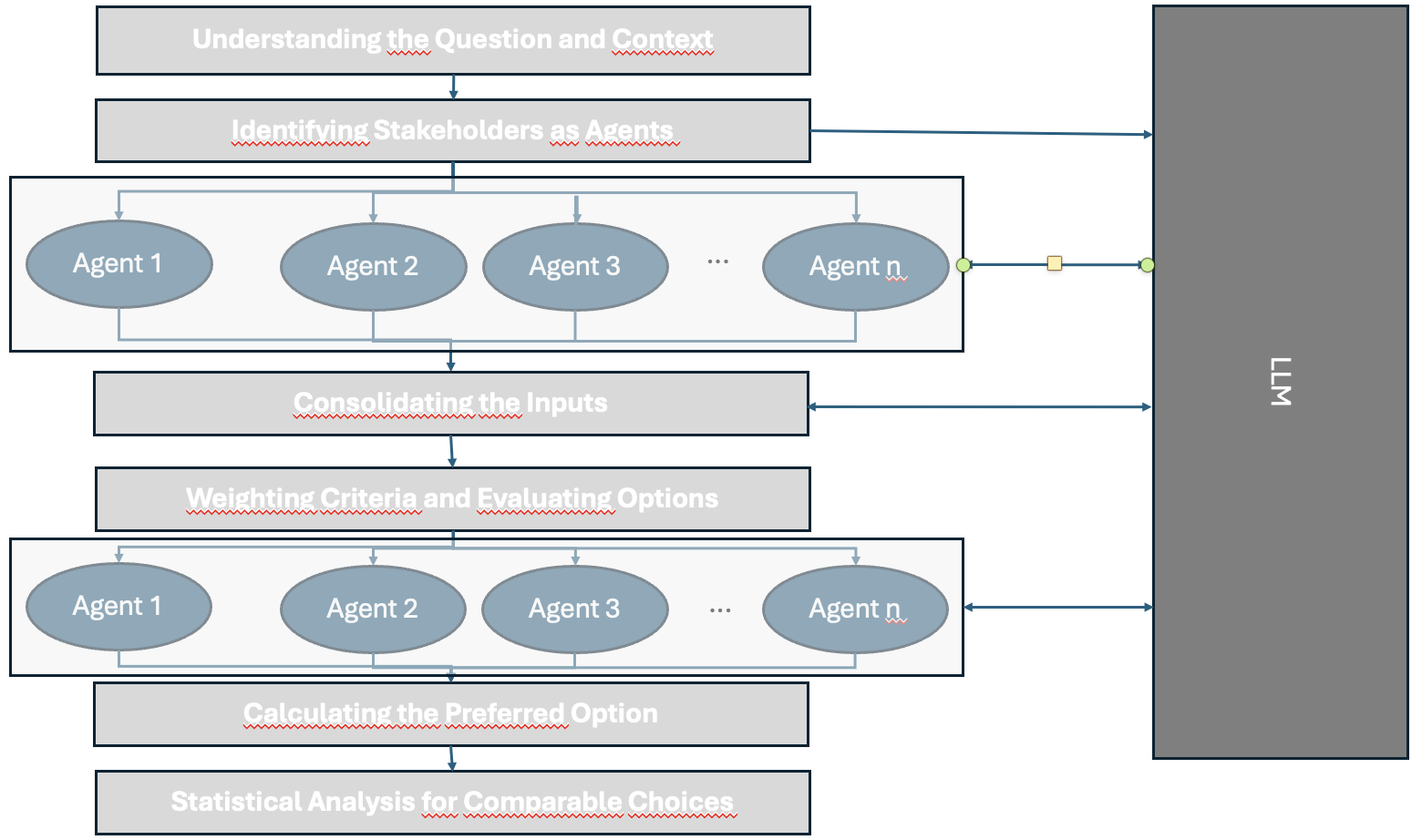}
\caption{An agent-based implementation of the QOC process.}
\label{fig:qoc_ai_agents}
\end{figure}

In an LLM-based implementation, these steps, including the identification of relevant options and criteria, as well as their corresponding evaluations, can be executed by a set of autonomous AI agents instantiated to represent the perspectives and priorities of different stakeholders involved in the decision (see Figure \ref{fig:qoc_ai_agents}). 
Before proceeding to the weighting and evaluation phase, however, the individually identified options and criteria from the various agents must first be consolidated into a unified set. This consolidation ensures that overlapping and redundant elements are harmonized, thereby establishing a coherent foundation for the subsequent quantitative evaluation.
Beyond the mere computation of a preferred option, the agent-generated data enable additional statistical analyses, such as the detection of outlier evaluations or divergences in the perceived importance of criteria or in how well the options support the criteria. Furthermore, transparency and explainability can be enhanced through meta-analysis of the decision process—for example, by identifying which criteria exerted the greatest influence on the final outcome.

\subsection{Sensitivity Analysis}
The sensitivity analysis process follows the structure of the Vester model but adapts it for LLM-driven automation and interpretability. It begins with an intake agent that transforms a natural-language scenario into a structured system definition, including the system’s goal, main subsystems, boundaries, and relevant stakeholders. Each stakeholder perspective, represented by an LLM adopting the persona of that stakeholder, is used to suggest candidate variables, ensuring that social, technical, and environmental aspects are captured.

Next, a cleaning and consolidation step merges overlapping or redundant variables while preserving those explicitly mentioned by the user. The outcome is a coherent set of system variables that form the foundation for the subsequent influence analysis.

An LLM-based evaluation module then estimates the direct influence of each variable on every other variable, producing a signed, weighted influence matrix. The numerical entries represent both the strength and direction of influence, indicating how changes in one factor impact another and whether the effect is positive or negative. This matrix acts as a transparent, data driven representation of the system’s internal dynamics.

From this matrix, the analyzer computes key statistical indicators to identify systemic roles:
\begin{itemize}
    \item \textbf{Active Sum (AS):} Measures how strongly a variable influences others. High values identify drivers and leverage points.
    \item \textbf{Passive Sum (PS):} Reflects how much a variable is affected by others, highlighting sensitive or dependent factors.
    \item \textbf{Role classification:} Variables are categorized into four standard roles: Act\-ive, Critical, Reactive, and Buffering—based on their relative AS and PS values.
\end{itemize}

These metrics reveal system dynamics, interdependencies, and potential feedback loops, forming the statistical backbone of the analysis.

An interpretation agent summarizes the results in natural language, outlining main drivers, dependencies, and leverage points. Visualizations, such as influence matrices or network diagrams, make causal links and systemic behavior transparent at a glance.

\subsection{Normal-Form Game (Game Theory)}
The normal-form game process is structured into clear, modular steps, each executed by a dedicated component or agent. First, a player-extraction module identifies the two players, their overall objectives, the payoff unit (for example, profit in USD, utility points, or another measurable outcome), and an expected value range derived from the user’s scenario description. Next, a strategy-generation module proposes up to ten strategies per player, ensuring that each is clearly named and accompanied by a short, descriptive summary of its intent.

Based on the extracted strategies, an LLM-powered payoff assignment module evaluates all possible strategy combinations and constructs the payoff matrix. This matrix forms the statistical foundation of the analysis, it captures the outcome values for every combination of player decisions. From this, the system can apply classical game-theoretic solution concepts that rely on these statistics to explain strategic behavior.

The deterministic analyzer computes three central measures:

\begin{itemize}
\item \textbf{Nash equilibria:} Stable states of the game where no player can gain by unilaterally changing their strategy.
\item \textbf{Pareto-optimal outcomes:} Solutions where no player’s payoff can be improved without reducing the other’s.
\end{itemize}

Finally, an interpretation agent summarizes results in natural language, explaining the equilibrium, key trade-offs, and strategy alignment with player goals.

\subsection{Sequential Game (Game Theory)}
The sequential-game pipeline extends the normal-form approach by incorporating the dimension of time and order into the decision process. It mirrors the same intake, identifying players, objectives, payoff units, and value ranges, but adds a layer of role conditioning. A role-assignment module creates concise persona prompts (e.g., “You are the company’s CEO deciding whether to expand into a new market”), which guide each player-specific LLM agent to reason consistently from its perspective.

Tree construction proceeds step-by-step from the root node, where players alternate in taking turns. At each node, the active player’s agent proposes one to three concrete actions. The reasoning at every decision point is grounded in the player’s role, not as a neutral observer. To ensure meaningful and interpretable trees, a configurable horizon policy limits the maximum depth, preventing unnecessarily long or repetitive paths while maintaining the completeness of possible outcomes.

Once the decision tree is built, each terminal node represents a possible end state of the scenario. A neutral evaluator agent then assigns numerical payoffs to both players, using the entire decision path and contextual information such as roles, objectives, and constraints. These payoffs are stored together with a short rationale at each terminal node, creating an auditable trail of how every outcome was scored.

The statistical information contained in the payoff structure allows a deterministic analyzer to compute key game-theoretic measures that explain the strategic dynamics of the sequential process:

\begin{itemize}
\item \textbf{Dominated paths:} Sequences of decisions that consistently lead to worse outcomes compared to alternatives, helping identify which strategies can be ruled out.
\item \textbf{Pareto-optimal paths:} Decision sequences that represent efficient trade-offs—no player’s result can improve without worsening the other’s.
\item \textbf{Subgame Perfect Equilibrium (SPE):} The most rational strategy profile at every stage of the game, determined via backward induction.
\end{itemize}

The analyzer outputs a concise summary of the equilibrium path, payoff structure, and player diagnostics. An interpretation agent translates these results into clear explanations, outlining the logic behind the SPE, key trade-offs, and recommendations aligned with user objectives. Visualizations such as annotated game trees and payoff charts further clarify decision flows and the rationale behind the strategies.

\subsection{Risk Management}
A risk management process can be effectively implemented using a multi-agent, LLM-based approach. Given a specific project or initiative, autonomous agents, each representing distinct stakeholder perspectives such as technical, organizational, financial, or legal viewpoints, can independently identify potential risks associated with the endeavor. The individually generated risk lists are then consolidated into a unified and non-redundant set of risks, ensuring that overlapping or synonymous risk descriptions are harmonized while maintaining the diversity of perspectives contributed by the agents.

After consolidation, each agent evaluates the identified risks along two principal dimensions: the probability that a risk will occur and the impact that its occurrence would have on the project’s objectives. These dimensions can be expressed using standardized scales (e.g., low–medium–high categories or numerical likelihood–impact matrices), allowing for quantitative aggregation and prioritization of risks. Based on these evaluations, composite risk scores can be computed to identify the most critical threats requiring mitigation or monitoring.

In this scenario, it may be advantageous to replace the general-purpose LLM with either a fine-tuned model or a Retrieval-Augmented Generation (RAG) architecture. Such adaptations allow the system to access and incorporate the most recent project documentation, technical reports, or relevant external sources, ensuring that the agents operate on an up-to-date and context-specific knowledge base. This integration significantly enhances the accuracy and relevance of the identified risks by grounding the agents’ assessments in the current state of the project and its environment.

Finally, statistical and analytical methods can be applied to the agents’ evaluations to detect inconsistencies, outliers, or systematic differences across stakeholder groups. These analyses improve the robustness and reliability of the overall assessment and contribute to greater transparency by revealing which risks are perceived as particularly significant and how alignment or divergence emerges among the participating agents.

\section{Evaluation}
This section presents an empirical evaluation of the proposed LLM-driven standard processes. To assess their validity and performance, selected implementations were tested on real-world datasets and reference scenarios. The results demonstrate the extent to which these AI-based processes can reproduce human-level reasoning and deliver interpretable, statistically verifiable outcomes.
\subsection{Evaluation of the QOC Approach}
To empirically assess the performance of the proposed AI-driven QOC process, an evaluation was conducted using real-world decision data from a Decentralized Autonomous Organization (DAO). A total of 102 investment proposals, including both the proposal content and the actual DAO voting outcomes, were collected via the DAO’s public API. Each proposal was then independently processed using the LLM-based QOC framework described in Section 4.1, and the AI-generated decisions were compared to those made by the DAO community.

For each model, a contingency table was constructed to record the frequency of agreement and disagreement between human (DAO) and AI decisions. Specifically, four categories were considered:
\begin{itemize}
\item DAO Y / AI Y — both DAO and AI approved the proposal,
\item DAO Y / AI N — DAO approved, AI rejected,
\item DAO N / AI Y — DAO rejected, AI approved,
\item DAO N / AI N — both DAO and AI rejected.
\end{itemize}
A statistical significance test (McNemar’s test) was applied to each contingency table to determine whether the differences between DAO and AI decisions were significant. The results are summarized in Table \ref{tab:qoc_eval}.
\begin{table}[htb]
\centering
\caption{Contingency table results for three LLMs compared to DAO decisions}
\label{tab:qoc_eval}
\begin{adjustbox}{width=\textwidth}
\begin{tabular}{lcccccc}
\toprule
\textbf{Model} & \textbf{DAO Y / AI Y} & \textbf{DAO Y / AI N} & \textbf{DAO N / AI Y} & \textbf{DAO N / AI N} & \textbf{Significance (p)} & \textbf{Cost} \\
\midrule
OSS-20b           & 70 (68.6\%) & 0 (0\%)     & 27 (26.5\%) & 5 (4.9\%)   & $2.03\times10^{-7}$ & 270 \\
OSS-120b          & 63 (61.8\%) & 7 (6.9\%)   & 24 (23.5\%) & 8 (7.8\%)   & 0.0023              & 247 \\
GPT-4-mini        & 56 (54.9\%) & 14 (13.7\%) & 20 (19.6\%) & 12 (11.8\%) & 0.303 (ns)          & 214 \\
Claude 4.5-Haiku  & 36 (35.2\%) & 34 (33.3\%) & 12 (11.8\%) & 20 (19.6\%) & 0.0012              & 154 \\
Claude 4.5-Sonnet & 36 (35.3\%) & 34 (33.3\%) & 10 (9.8\%)  & 22 (21.6\%) & 0.0003              & 134 \\
GPT-5-mini        & 32 (31.4\%) & 38 (37.3\%) & 8 (7.8\%)   & 24 (23.5\%) & $7.2\times10^{-7}$  & 118 \\
GPT-5             & 24 (23.5\%) & 46 (45.1\%) & 3 (2.9\%)   & 29 (28.4\%) & $8.1\times10^{-10}$ & 76 \\
\bottomrule
\end{tabular}
\end{adjustbox}
\end{table}
To complement the statistical comparison, a cost function was introduced to capture the relative severity of decision mismatches between the DAO and the AI models:
\begin{center}
$c = 1 * d + 10 * a$    
\end{center}
where
\begin{itemize}
    \item c denotes the total cost,
    \item d is the number of false negatives (proposals accepted by the DAO but rejected by the AI),
    \item a is the number of false positives (proposals rejected by the DAO but accepted by the AI).
\end{itemize}
This cost function assumes that false positives are substantially more detrimental, representing a potential misallocation of DAO funds, than false negatives, which merely reflect missed opportunities, but increases the aspect of funds-protection.

The analysis reveals clear differences in decision behavior between the DAO and the AI-driven QOC approach. Most models exhibited statistically significant deviations from the DAO’s decisions, with GPT-4-mini being the only model for which the difference was not significant.

The smaller, open-weight models (OSS-20b and OSS-120b) showed a strong tendency to approve proposals that the DAO had rejected, reflected in high false-positive rates and correspondingly high costs. This behavior suggests that such models exhibit insufficient conservatism for use in financial or governance-sensitive contexts.

In contrast, the more advanced LLMs (Claude 4.5 and GPT-5 families) displayed the opposite pattern, adopting a more conservative stance, rejecting a higher proportion of proposals that had been approved by the DAO. This behavior, while stricter, led to substantially lower cost scores, particularly in the case of GPT-5, which achieved the lowest overall cost while maintaining only three false positives.

Overall, the results suggest that larger, more capable LLMs tend to produce more conservative and risk-aware decisions within the QOC framework. 

\subsection{Evaluation of the Sensitivity Analysis}

We evaluate whether a multi-agent LLM pipeline can replicate the structure and reasoning of a Vester-style sensitivity analysis using IVL’s logistics study on apparel flows from China to the Nordics \cite{wolf2012logistic}. Because IVL later adds company-internal factors, we assess both the \emph{core transport system} and \emph{company/internal} layers.
The pipeline is executed 100 times, prompted only with a concise transport-scope statement (no IVL variables disclosed). Each run outputs factors with descriptions, a signed influence matrix, Vester metrics (AS, PS, $P = \mathrm{AS} \cdot \mathrm{PS}$, $Q = \mathrm{AS}/\mathrm{PS}$) with role labels, loops, and a short interpretation.
To measure semantic correspondence, sentence embeddings (all-mpnet-base-v2) are used to match LLM factors to IVL factors via cosine similarity ($\geq 0.5$), a paraphrase-level threshold consistent with prior work \cite{corley_mihalcea_2005}. We then assess factor alignment and role consistency across both strata. Additionally, an LLM judge (GPT-5 Thinking), following evidence that LLM judgments correlate well with humans \cite{zheng2023judging},  rates each study and a reconstructed IVL baseline on an eight-criterion rubric (max 100) covering boundary, stakeholders, variables, causality, loops, scenarios, temporal scope, and actionability.

Across all \textbf{26} IVL factors, mean alignment is \textbf{55.5\%} (range \textbf{23.1--80.8\%}); \textbf{71\%} of runs reach $\geq 50\%$. 
For the \emph{core} 18 factors, alignment rises to \textbf{62.9\%} (range \textbf{27.8--94.4\%}), while the \emph{internal/company} 8 average \textbf{38.8\%}. 
High matches occur for \emph{Fuel prices} (99\%), \emph{Competition in the transport market} (95\%), and \emph{Environmental demands} (93\%), whereas structural items (e.g., \emph{Total amount of goods}) and abstract internal variables (e.g., \emph{Image of company}) align less. 
Role fidelity averages \textbf{56.6\%}, reflecting typical volatility in AS/PS-based classifications.
Rubric scores are tightly clustered (\textbf{77--98}, mean \textbf{92.97}, SD \textbf{3.47}), nearly identical to IVL's \textbf{93}, with consistently high marks for boundary (\textbf{4.99/5}), actionability (\textbf{14.9/15}), and temporal reasoning (\textbf{13.5/14}). 
This indicates that even with minimal prompting, the pipeline reproduces the methodological form of a Vester analysis: coherent factorization, causal mapping, and loop construction rather than textual imitation. 
Overall, the results show that a conditioned multi-agent LLM can operationalize the Vester method and achieve human-level coherence in structure and interpretation. Remaining gaps cluster around internal company factors and role stability, precisely the aspects a \emph{specialist layer} could improve by grounding analyses in richer, domain-specific context.

\subsection{Evaluation of the Game Theory - Marcel}

We test whether a role-conditioned, LLM-driven extensive-form game can reproduce the strategic dynamics and end-state logic of the Cuban Missile Crisis abstraction \cite{zagare2014cuban}, aligning its SPE path with the ground-truth labeled sequence: 
(1) limited coercion (quarantine/interdiction with public warning, \textsc{Escalation}); 
(2) public pushback with private willingness to negotiate (backchannels, \textsc{Signal/Info}); 
(3) assurance exchange (non-invasion pledge and Jupiter context, \textsc{De\mbox{-}escalation}); and 
(4) withdrawal and dismantlement with verification (\textsc{De\mbox{-}escalation}).

We execute $60$ independent runs with a fixed prompt and depth cap $D{=}4$ (max four moves/path). At each node, the active player's persona proposes concrete actions; payoffs are then evaluated and backward induction selects the SPE path. 
To evaluate each step, we label each step using a compact taxonomy Escalate, Signal/Info, De-escalate, Wait under clear rules. Labeling is performed by an LLM annotator (GPT-5 Thinking) given the same codebook for both model outputs and the paper-derived reference, consistent with evidence that GPT-class models can match human annotation quality under structured instructions \cite{gilardi2023chatgpt}. 
Some SPE paths terminate after three steps while others reach four steps: Three-step paths (17/60) are compared to paper steps 1--3; four-step paths (43/60) to steps 1--4. In addition to stepwise alignment, we compute a path-level label.

For three-step runs ($n{=}17$), match rates vs.\ paper labels at positions 1--3 are \textbf{0.529}, \textbf{0.294}, \textbf{0.647}. For four-step runs ($n{=}43$), match rates at positions 1--4 are \textbf{0.535}, \textbf{0.442}, \textbf{0.349}, \textbf{1.000}. Thus, openings align in roughly half the cases, mid-sequence alignment is mixed (more variance at steps 2--3), and the terminal decision matches \textsc{De-escalate} in all four-step runs.

Across all runs, the path-level label equals \textsc{De-esca\-late} in \textbf{56/60} cases (\textbf{93.3\%}); by length: \textbf{15/17} (88.2\%) for three-step and \textbf{41/43} (95.3\%) for four-step paths.

The role-conditioned, LLM-driven extensive-form game reliably reproduces the \emph{end-state logic} of the Cuban Missile Crisis abstraction \cite{zagare2014cuban}, with \textbf{93\%} of all paths and \textbf{100\%} of four-step paths ending in \textsc{De\mbox{-}escalate}. 
While mid-sequence variance appears in the transitions, this reflects the presence of multiple continuations under minimal prompting. 
A \emph{specialist layer} could reduce such variance by injecting structured, context-rich knowledge, e.g., historical documents, strategic playbooks, or actor-specific constraints, to guide belief updates and action generation. 
Overall, the model not only reproduces the historical outcome but also mirrors the \emph{patterned reasoning dynamics} characteristic of sequential-game processes.

\section{Outlook and Future Work}
Despite these promising outcomes, several limitations and potential risks must be acknowledged. First, retrieval-augmented generation (RAG) pipelines, although essential for domain grounding, may inadvertently propagate outdated, incomplete, or biased information from external sources. Without robust source validation and temporal freshness checks, these retrieval errors can compromise both the factual accuracy and the traceability of the final explanations.

Second, hallucination propagation remains a challenge: when agents generate fabricated or inconsistent data, such artifacts can cascade through the reasoning chain, ultimately distorting aggregate results. While the proposed layered design and explainability barrier help to isolate these effects within controlled process layers, additional consistency-checking and reference validation mechanisms are required to ensure epistemic reliability in multi-step analyses.

Third, misaligned agent outputs can occur in multi-agent settings, where different role-conditioned agents interpret prompts or contextual cues divergently. In the presented architecture, this risk is mitigated through statistical outlier detection, consensus formation, and aggregation procedures that treat extreme or incoherent agent responses as diagnostic rather than determinative signals. Nonetheless, future work should formalize these reconciliation mechanisms, e.g. , by integrating alignment scoring or trust-weighting models that dynamically adjust each agent’s influence according to its historical reliability and contextual fit.

Finally, although LLM-driven standard processes improve interpretability, they still depend on the opaque latent reasoning of the underlying model. The approach therefore cannot guarantee full transparency at the cognitive level of the LLM; it can only expose structured traces above the explainability barrier. This partial transparency should be recognized as an inherent limitation of current generative architectures.

Future research will focus on technical and methodological refinements addressing the above limitations. A first priority is the implementation of an integrated, modular software environment in which specialized agents operate under explicit retrieval, grounding, and validation constraints. RAG pipelines will be equipped with source-credibility filters and timestamp controls to reduce hallucination and data drift.

A second focus lies in formal process orchestration, developing a routing layer that interprets user intent, selects the appropriate analytical module (e.g., QOC vs. Game Theory), and ensures transparent data flow between agents. Complementary schema formalization, defining inputs, evaluation functions, and output structures in a machine-readable format, will further enhance reproducibility and auditability.

In the longer term, process-aware LLMs represent a compelling direction. By fine-tuning models on structured reasoning patterns, decision-analytic schemas could become part of the model’s intrinsic latent space, reducing reliance on external orchestration and improving internal alignment. Combined with adaptive outlier-management and grounding protocols, this evolution would yield AI systems that are not only explainable in their outputs but also explainable by design in their reasoning.

\end{document}